%% file: main.tex
\documentclass{article}
\usepackage{spconf,amsmath,graphicx}
\usepackage[utf8]{inputenc}
\usepackage{makecell}
\usepackage{color}
\usepackage{array,multirow}
\usepackage{siunitx,booktabs}
\usepackage{comment}
\usepackage{microtype}
\usepackage{listings}
\usepackage{caption}
\title{Input Dropout for Spatially Aligned Modalities}
\name{Sébastien de Blois, Mathieu Garon, Christian Gagné\sthanks{Canada-CIFAR AI Chair}, Jean-François Lalonde }
\address{Université Laval}

\newcommand\mc[1]{\multicolumn{1}{c}{#1}}

\usepackage{xcolor}

\newcommand{\sizeB}{2.2}

\begin{document}
%
\maketitle
\begin{abstract}

Computer vision datasets containing multiple modalities such as color, depth, and thermal properties are now commonly accessible and useful for solving a wide array of challenging tasks. However, deploying multi-sensor heads is not possible in many scenarios. As such many practical solutions tend to be based on simpler sensors, mostly for cost, simplicity and robustness considerations. 
In this work, we propose a training methodology to take advantage of these additional modalities available in datasets, even if they are not available at test time. By assuming that the modalities have a strong spatial correlation, we propose \emph{Input Dropout}, a simple technique that consists in stochastic hiding of one or many input modalities at training time, while using only the canonical (e.g. RGB) modalities at test time. We demonstrate that \emph{Input Dropout} trivially combines with existing deep convolutional architectures, and improves their performance on a wide range of computer vision tasks such as dehazing, 6-DOF object tracking, pedestrian detection and object classification. 
\end{abstract}

\input{intro_icip}

\input{method_icip}

\input{dehazing_icip}
\input{clf_icip}
\input{others_icip}

\input{discussion_icip}

\input{ack}

{
\bibliographystyle{ieee}
\bibliography{refs}
}

\end{document}

%% file: intro_icip.tex
\section{Introduction}

The use of deeper networks and data-hungry algorithms to solve challenging computer vision problems has created the need for ever richer datasets. In addition to common image datasets such as the famed ImageNet~\cite{deng2009imagenet}, datasets containing multiple modalities have also been collected to address a variety of problems ranging from depth estimation~\cite{Silberman:ECCV12}, indoor scene understanding~\cite{xiao2013sun3d}, 6-DOF tracking~\cite{garon2018framework}, multispectral object detection~\cite{hwang2015multispectral}, autonomous driving~\cite{Geiger2012CVPR,Cordts2016Cityscapes} to haze removal~\cite{sakaridis2018semantic}, to name just a few. More generally, learning from multiple modalities has been explored to determine which ones are useful~\cite{wu2004optimal}, and multiple ways of combining them have been proposed~\cite{gunes2005affect, hu2015jointly, mees2016choosing, wagner2016multispectral}.



\begin{figure}
\captionsetup{font=small}
\small
    \includegraphics[width=\linewidth]{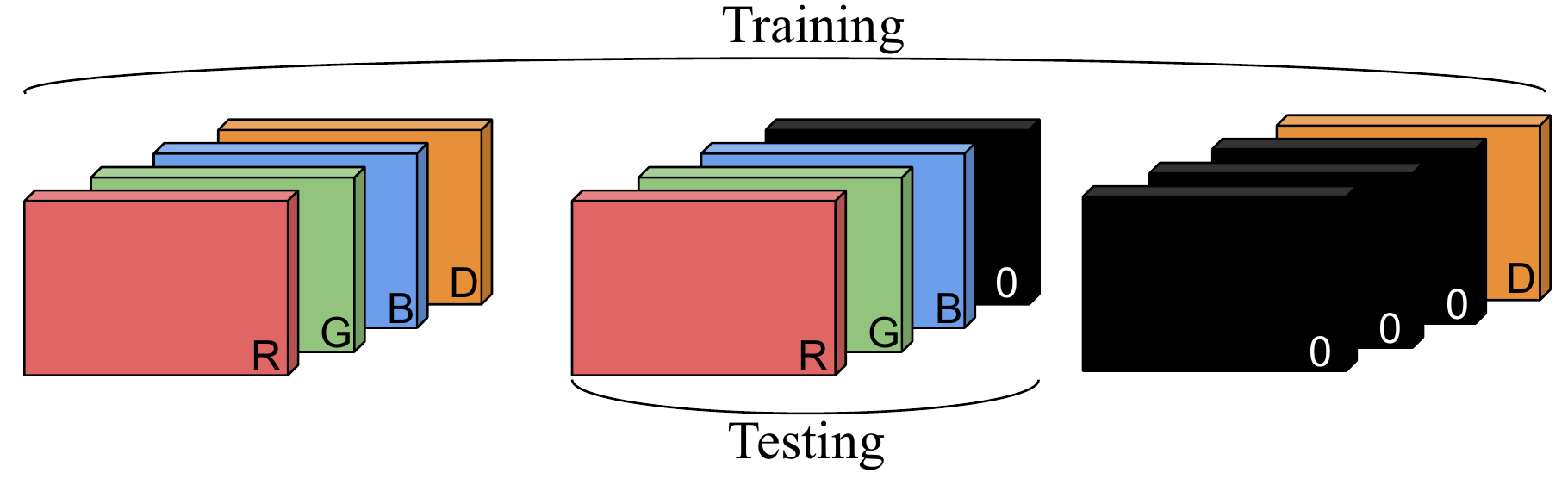}
    \caption{The Input Dropout strategy for a given RGB image and additional modality (e.g.~depth in orange). At training time, the additional modality is concatenated to the RGB image, with a given probability the RGB modality or the additional modality is being set to 0 (black). At test time (middle), the additional modality is always unavailable (i.e., set to 0). In training, for the \emph{addit} mode, only the two left cases are used, while for the  \emph{both} mode, all three cases are used.}
    \label{fig:architecture-inputdrop}
\end{figure}

Training deep learning models on additional modalities typically means that these extra modalities must also be available at test time. Unfortunately, capturing more modalities requires significant time and effort. Adding sensors alongside an RGB camera results in increased power consumption, less portable setups, the need to carefully calibrate and synchronize each sensor, as well as additional constraints on bandwidth and storage requirements. This may not be practical for multiple applications---including augmented reality, robotics, wearable and mobile computing, etc.---where these physical constraints preclude the use of additional sensors.

This dichotomy between the advantage brought by additional modalities and the impediment they impose on real systems has attracted attention in the literature. Can we train on additional modalities without relying on them at test time? In their ``learning with privileged information'' paper, Vapnik et al.~\cite{vapnik2009new} introduce a theoretical framework which shows that this may indeed be feasible. Practical techniques have since been introduced, but those tend to be specifically targeted towards specific network architectures and applications.  For example, ``modality hallucination''~\cite{hoffman2016learning} and variants~\cite{garcia2018modality,garcia2019learning} propose to train networks on different modalities independently, and shows that by changing the input modality of one of the networks while forcing the latent space to keep its former structure improves convergence. In \cite{neverova2016moddrop}, authors proposed to independently process multiple modalities in parallel branches within a network, and fuse the resulting feature maps using so-called ``modality dropout'' to make the network invariant to missing modalities. Despite improving performance, these methods are complex to implement, may require multiple training steps, and must be adapted differently to each problem.

In this paper, we propose a technique for exploiting additional modalities at training time, without having to rely on them at test time. In contrast to previous work which requires task-specific architectures~\cite{neverova2016moddrop} or multiple training passes~\cite{garcia2019learning}, our approach is extremely simple (can be implemented in a few lines of code), is independent of the learning architecture used, and does not require any additional training pass. Assuming that modalities are spatially-aligned and share the same spatial resolution, we propose to randomly dropout~\cite{JMLR:v15:srivastava14a} entire input modalities at training time. At test time, the missing modality is simply set to 0. We demonstrate that our proposed strategy, \emph{Input Dropout}, can be leveraged to obtain between 2--20\% gain over training on RGB-only, on a variety of applications. 



%% file: method_icip.tex
\section{Input Dropout}

\paragraph*{Assumptions}

We assume that all input modalities are spatially aligned and can be represented as additional channels of the same input image. In our experiments, we also assume that the RGB modality is the only modality available at test time, therefore the other modality is never available during testing.

\paragraph*{Approach}

Our proposed \emph{Input Dropout} strategy is illustrated in fig.~\ref{fig:architecture-inputdrop}. The additional modality is first channel-wise concatenated to the RGB image, and the resulting tensor is fed as input to the neural network. The first convolutional layer of the network must be adapted to this new input dimensionality (c.f. sec.~\ref{sec:classification}). At training time, one of the input modalities is randomly set to 0 with probability $P_\mathrm{drop} \in [0,\,1]$. This effectively ``drops out''~\cite{JMLR:v15:srivastava14a} the corresponding modality. At test time, the additional modality is always set to 0. Implementing \emph{Input Dropout} requires a few lines of PyTorch code. 

Since we assume a single additional modality is combined with an RGB image, we are faced with two options. We could randomly drop only the additional modality and always keep the RGB (we dub this option \emph{addit}), or drop either the RGB or the additional modality (\emph{both}). In these two cases, a uniform probability distribution for the different possible cases is used. For the \emph{addit} mode, the probability of dropping the additional modality is set to $P_\mathrm{drop} = 0.5$. For the \emph{both} mode, the probability of dropping either the RGB or the additional modality is $P_\mathrm{drop} = 0.33$.

Our method is mainly related to ``modality dropout''~\cite{neverova2016moddrop}, which fuses the modalities in a learned latent space. Their main limitation is that specialized network branches must be learned for each modality, which adds complexity. In contrast, our method can be used on existing convolutional architectures with very little change. We will compare to \cite{neverova2016moddrop} in sec.~\ref{sec:classification}.

%% file: dehazing_icip.tex
\section{Input Dropout for image dehazing}

We first experiment with \emph{Input Dropout} on single image dehazing~\cite{Li_2018_CVPR,yang2018towards} with depth (RGB+D) as the additional modality available at training time only. For this, we employ the D-Hazy dataset~\cite{ancuti2018ntire}, which contains 1449 pairs of RGB+D images where haze is synthetically added on images from the NYU Depth dataset~\cite{Silberman:ECCV12}. We use 1180 images in training, 69 for validation, and 200 for test. Our model is similar to \cite{Li_2018_CVPR}, the only difference being that the generator is a ResNet (with nine blocks) as in \cite{johnson2016perceptual}.

Similar to~\cite{Li_2018_CVPR,yang2018towards}, the network is trained on a combination of a GAN, a pixel-wise $\mathcal{L}_1$, and a perceptual loss~\cite{johnson2016perceptual} to preserve the sharpness of the image:
\begin{equation}
\mathcal{L}_\mathrm{generator}=\mathcal{L}_\mathrm{GAN}+\lambda_{1}\,\mathcal{L}_\mathrm{1}+\lambda_{2}\,\mathcal{L}_\mathrm{percep} \,,
\label{eq:h1}
\end{equation}
where $\lambda_{1} = \lambda_{2} = 10$ (obtained with grid search on the validation set). At training time, \emph{Input Dropout} uses the \emph{addit} mode. Indeed, it does not make sense to drop the RGB image since it would be equivalent to obtain a haze-free image from a depth map. We also experiment on single image dehazing using segmentation (RGB+S) as an additional training modality.The Foggy Cityscape Dataset~\cite{sakaridis2018semantic}, an extension of Cityscapes~\cite{Cordts2016Cityscapes} which contains ground truth scene segmentations is used here. The same network and training procedure are used. 

\begin{figure*}[t] 
\captionsetup{font=small}
  \small
  \centering
  \begin{tabular}{ccccc}

  \includegraphics[width=0.16\linewidth]{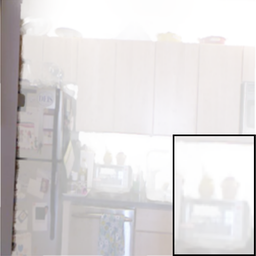} & 
  \includegraphics[width=0.16\linewidth]{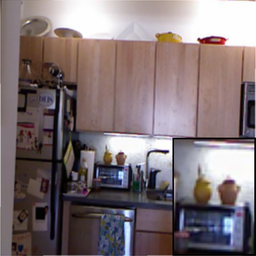} & 
  \includegraphics[width=0.16\linewidth]{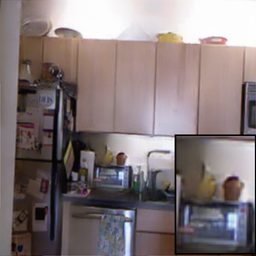} & 
  \includegraphics[width=0.16\linewidth]{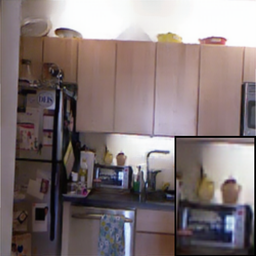} & 
  \includegraphics[width=0.16\linewidth]{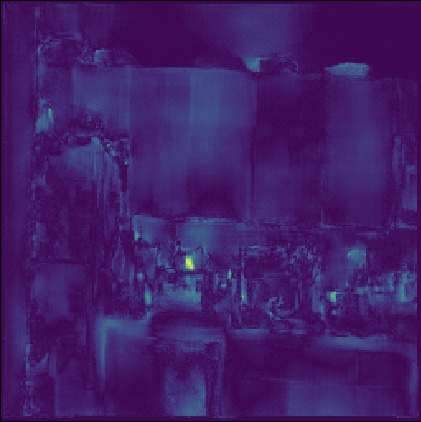}  \\

  \includegraphics[width=0.16\linewidth]{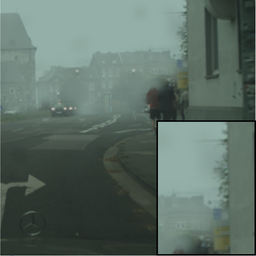} & 
  \includegraphics[width=0.16\linewidth]{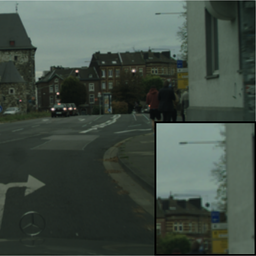} & 
  \includegraphics[width=0.16\linewidth]{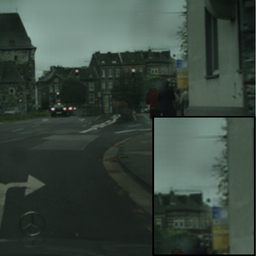} & 
  \includegraphics[width=0.16\linewidth]{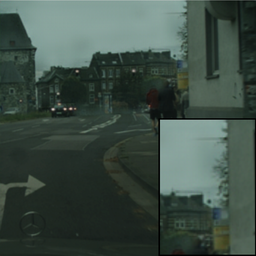} & 
  \includegraphics[width=0.16\linewidth]{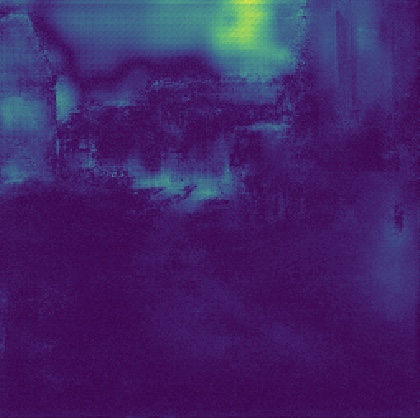} \\
  
  Input & Ground truth & RGB only & \emph{Input Dropout} & Difference
  \end{tabular}
    
  \caption{Qualitative examples for dehazing RGB images from (top row) D-Hazy~\cite{ancuti2016d} and (bottom row) Foggy Cityscapes~\cite{sakaridis2018semantic,Cordts2016Cityscapes}. From left to right: hazy input, ground truth haze-free image, results when trained on RGB only, results with \emph{Input Dropout}, absolute difference between the 3rd and 4th column, shown using a color map ranging from blue (low) to yellow (high).}
  \label{fig:RGBD-dehazing}
\end{figure*}

Quantitative dehazing results with \emph{Input Dropout} are provided in tab.~\ref{table:RGBD-dehazing}, and corresponding representative qualitative results in fig.~\ref{fig:RGBD-dehazing}. Over the RGB-only baseline, relative improvements of 3.6\% and 3.4\% on PSNR and SSIM respectively are observed when using \emph{Input Dropout} on RGB+D, and 4.5\% PSNR and 2.2\% SSIM for RGB+S. We also compare our method to competing techniques, such as ``Dehazing for segmentation'' (D4S)~\cite{deBl1909:Learning} which proposes an approach to dehaze to increase performance for a subsequent task using a  modality only available during training, and Pix2Pix GAN~\cite{bischke2018overcoming} which employs an extra generator to generate the missing modality from the RGB image. In every case, \emph{Input Dropout} performs better while being simpler than the other approaches. Note that we have not compared our approach to ``modality distillation''~\cite{garcia2019learning} here since the method cannot be applied to this scenario. Indeed, it would involve training a network to dehaze a depth (or segmentation) image, which would require hallucinating scene contents. 






\begin{table}[t]
\captionsetup{font=small}
    \small
    \centering
    \begin{tabular}{ccccc} 
    \toprule
    & \multicolumn{2}{c}{RGB+D} & \multicolumn{2}{c}{RGB+S} \\
        Methods                 & PSNR    &  SSIM  & PSNR   &  SSIM \\
    \midrule
      RGB-only      & 17.61             & 0.74        & 23.55 & 0.91 \\
      D4S \cite{deBl1909:Learning}            & N/A & N/A         & 23.90 & 0.92 \\
      Pix2Pix GAN \cite{bischke2018overcoming}         & 17.70        & 0.75        & 22.90              & 0.91 \\
      \emph{Input Dropout}          & \textbf{18.24}           & \textbf{0.76}    & \textbf{24.60}& \textbf{0.93}    \\ 
        \bottomrule
    \end{tabular}
\caption{Quantitative results for single image dehazing using an additional depth (RGB+D) and segmentation (RGB+S) modality at training time. Results are reported on the D-Hazy dataset~\cite{ancuti2016d} for RGB+D and the Foggy Cityscapes dataset~\cite{sakaridis2018semantic} for RGB+S. For each technique, the average over five different training runs are reported. In all scenarios, \emph{Input Dropout}, despite its simplicity, is the technique that provides the largest improvement over the RGB-only baseline. RGB+D: Statistically significant results are in bold, Wilcoxon–Mann–Whitney (WMW) test with an $\alpha$ of 0.025. RGB+S: Statistically significant results are in bold, WMW test with an $\alpha$ of 0.05 for PSNR and 0.025 for SSIM.}
    \label{table:RGBD-dehazing}
\end{table}

%% file: clf_icip.tex
\section{Input dropout for classification}
\label{sec:classification}

We evaluate the use of \emph{Input Dropout} for image classification using RGB+D training data. For this, we rely on the methodology proposed by Garcia et al.~\cite{garcia2019learning}, who use the crops of individual objects from the NYU V2 dataset \cite{Silberman:ECCV12} adapted by \cite{hoffman2016learning} for object classification using RGB+D. We used the same split as in~\cite{garcia2019learning}: 4,600 RGB-D images in total, where around 50\% are used for training and the remainder for testing. Here, we rely on a ResNet-34~\cite{he2016deep}, initialized with pretrained weights on ImageNet~\cite{deng2009imagenet}. To adapt the pretrained ResNet-34 to use \emph{Input Dropout}, we append additional channels to the filters of the first convolution layer and initialize the new weights randomly. Doing so preserves the pretrained weights for the RGB modality. 

Tab.~\ref{tab:IPvsMD_PT} shows the quantitative classification accuracy obtained with the various methods. First, we report results when the depth modality is available at test time to provide an upper bound on performance. Next, we evaluate training a single network on the RGB modality only (``RGB-only''), the approach of \cite{bischke2018overcoming} which relies on a GAN to hallucinate the depth at test time, and our \emph{Input Dropout} strategy (in the \emph{addit} mode). Our approach provides the best results, despite being the simplest.

We further compare to ensemble methods. First, two networks trained on RGB only, with their answers averaged before the argmax, yield an absolute performance improvement of 3.2\% over the single-network baseline. The ``modality distillation'' approach of Garcia et al.~\cite{garcia2019learning} relies on a combination of two networks: one trained on RGB only, and another, so-called ``hallucination'' network. That second network is trained to produce a latent representation that is similar to a proxy network trained on the depth modality only. The final output is the mean of the RGB-only and the hallucination network. We reimplemented their approach in PyTorch to ensure direct comparison with our results, which yields a 1.8\% absolute improvement over the RGB+RGB baseline. 

We directly compare our technique to ``modality distillation''~\cite{garcia2019learning} by using one network trained on RGB only, and another network trained on RGB+D with \emph{Input Dropout} (instead of their ``hallucination'' network). This yields approximately the same performance as ``modality distillation''~\cite{garcia2019learning}, despite being simpler to train, requiring a single network architecture and a single training pass (i.e. both networks can be trained in parallel, while they must be trained sequentially for \cite{garcia2019learning}).

\begin{table}[ht]
\captionsetup{font=small}
    \small
    \centering
    \begin{tabular}{ccc}
    
    \toprule
    Method & Ensemble & Accuracy \\
    \midrule

    RGB+D & No  & 58.9\%  \\
    Depth (D) only & No & 57.0\% \\
    \midrule
    RGB only & No  & 47.5\%  \\
    Pix2Pix GAN \cite{bischke2018overcoming} & No    & 48.2\% \\
    ModDrop \cite{neverova2016moddrop} & No & 44.3\% \\
    \emph{Input Dropout}  & No & \textbf{49.5\%} \\
    \midrule
    RGB+RGB & Yes  & 50.7\% \\
    Mod. distillation \cite{garcia2019learning} & Yes & \textbf{52.5\%} \\
    \emph{Input Dropout}  + RGB & Yes  & \textbf{52.7\%} \\
    \bottomrule

    \end{tabular}
    \caption{Classification accuracies on the NYU V2 dataset adapted by \cite{hoffman2016learning}. Results are the average over five different training runs. Statistically significant results are in bold, WMW test with an $\alpha$ of 0.05 for no ensemble and 0.025 for ensemble.}
    \label{tab:IPvsMD_PT}
\end{table}


%% file: others_icip.tex
\section{Other applications}
We evaluate \emph{Input Dropout} (in the \emph{both} mode) on two additional applications: tracking in RGB+D and pedestrian detection in RGB+thermal.

\subsection{3D object tracking with RGB+D}

We first focus on the problem of tracking 3D objects in 6 degrees of freedom (DOF). To do so, we employ the methodology of Garon et al.~\cite{garon2018framework}, who presented a technique for tracking a known 3D object in real-time using synthetic RGB+D data. They also provide an evaluation dataset containing 297 real sequences captured with a Kinect V2 with ground truth annotations of the 6-DOF poses of 11 different objects. 

Here, we focus on the ``occlusion'' scenario proposed by~\cite{garon2018framework} where the objects are rotated on a turntable while being partially hidden by a planar occluder with (measured) occlusion varying from 0\% to 75\%. We evaluate \emph{Input Dropout} using the same CNN architecture as in~\cite{garon2018framework}. For a given pose $\mathbf{P} = [\mathbf{R} \quad \mathbf{t}]$ where $\mathbf{t}$ is the translation and $\mathbf{R}$ the rotation matrix, the translation error is defined by its $L2$ norm and the rotation matrix distance is computed with:
\begin{equation}
\delta_R(R_1, R_2) = \arccos(\frac{\mathrm{Tr}(\mathbf{R}^{T}_{1}\mathbf{R}_2) - 1}{2}) \,,
\end{equation}
where $\mathrm{Tr}$ is the matrix trace~\cite{garon2018framework}.

Quantitative 6-DOF tracking results are reported in tab.~\ref{tab:deeptrack_eccv18_error}. We observe that \emph{Input Dropout} generally improves the results for the tracking task in translation with a relative gain as high as 33.3\% in the hardest sequences, and an average of 17\% relative gain in rotation. The error reported is the average of 5 training runs for each method. 

\begin{table}[t!]
\captionsetup{font=small}
    \centering
    \small
    \begin{tabular}{lccccc}
    	\toprule
    	& \multicolumn{2}{c}{Translation (mm)} & & \multicolumn{2}{c}{Rotation (degrees)} \\
    	\midrule
    	Occlusion \% & 0--30 & 45--75 & & 0--30 & 45--75	\\
        \midrule
    	RGB-only  & \textbf{22.3} & 43.2 & & 10.2 & 24.6 \\
    	\emph{Input Dropout}  & 22.8 & \textbf{28.8} & & \textbf{8.1} & \textbf{21.3} \\
    	Relative gain & -2.2\% & 33.3\% & & 20.5\% & 13.4\% \\
        \bottomrule
	\end{tabular}
	\caption{Tracking error in translation and rotation with respect to the ratio of occlusion from the dataset of Garon et al.~\cite{garon2018framework}. We observe that \emph{Input Dropout}  augment most of the scenarios significantly. In translation, the error with Input Dropout stabilizes after 45\% occlusion, and the average relative gain in rotation is 16.7\%.}
	\label{tab:deeptrack_eccv18_error}
\end{table}

\subsection{Pedestrian detection with RGB+T}

We experiment with pedestrian detection on RGB+T (thermal) images using the KAIST Multispectral pedestrian dataset~\cite{hwang2015multispectral}. The training/validation/test sets are composed of 16,000/1,100/3,500 pairs of thermal/visible images for nighttime and 32,000/1,500/8,500 for daytime. 

Here, we rely on RetinaNet~\cite{lin2017focal}, which is a state-of-the-art architecture for object detection. The RetinaNet is trained with a focal loss using a ResNet-34~\cite{he2016deep} and a Feature Pyramid Network (FPN)~\cite{lin2017feature} as backbone for feature extraction. The RetinaNet is initialized with pretrained weights on ImageNet~\cite{deng2009imagenet}. As in sec.~\ref{sec:classification}, additional channels are appended to the filters of the first convolutional layer to preserve the learned weights on RGB. The network is then fine-tuned on the KAIST images until convergence on the validation set. 

To evaluate performance, we compute the mean average precision (mAP) with an intersection-over-union (IoU) score of 0.5. Tab.~\ref{tab:RGBT-pretrained} shows the results of the experiments in both night- and daytime scenarios. We observe that our \emph{Input Dropout} strategy yields improvements in all cases, nighttime RGB pedestrian detection improves by 18.9\%, and daytime RGB pedestrian detection improves by 15.1\%.

\begin{table}[t!]
\captionsetup{font=small}
    \small
    \centering
    \begin{tabular}{ccc}        
    \toprule
                         
           Method   & Nighttime  & Daytime \\
    \midrule
     
        RGB-only    & 0.228  & 0.351  \\
        \emph{Input Dropout}   & \textbf{0.271}  & \textbf{0.404}\\
        Relative gain  & 18.9\% & 15.1\% \\
    \bottomrule
    \end{tabular}
    \caption{Mean average precision (mAP) with an IoU of 0.5 results with RGB+T for nighttime and daytime pedestrian detection with and without \emph{Input Dropout}, RGB only in test time. Each results column indicates the modality that is used at test time. The RGB-only row trains on the test modality only, while \emph{Input Dropout} uses both modalities at training time. Results are the average over five different training runs. The last row is the relative performance gain resulting from using \emph{Input Dropout}. Statistically significant results are in bold, WMW test with an $\alpha$ of 0.01 for nighttime and daytime.}
    \label{tab:RGBT-pretrained}
\end{table}

%% file: discussion_icip.tex
\section{Discussion}

We propose \textit{Input Dropout} as a simple and effective strategy for leveraging additional modalities at training time which are not available at test time. We extensively test our technique in several applications---including single image dehazing, object classification, 3D object tracking, and object detection---on several additional modalities---including depth, segmentation maps, and thermal images. In all cases, using \emph{Input Dropout} in training yields improved performance at test time, even if the additional modality is unavailable. Our approach, which can be implemented in a few lines of code only, can be used as a drop-in replacement with no change to the network architecture, aside from the addition of one extra input dimension to the first layer filters.
The main limitation of our approach is that we have experimented on adding only a single additional modality to the RGB baseline. In the future, we plan on exploring the applicability of the approach with more modalities. 





%% file: ack.tex
\section{Acknowledgements}
This research was supported by Thales, NSERC and the \mbox{NSERC}/Creaform Industrial Research Chair on 3D Scanning: CREATION 3D. We thank Nvidia Corporation for the donation of the GPUs used in this research.